\documentclass[10pt,twocolumn,letterpaper]{article}

\usepackage{cvpr}              

\usepackage[dvipsnames]{xcolor}

\usepackage{algorithm}
\usepackage{algpseudocode}
\usepackage{multirow}
\definecolor{cvprblue}{rgb}{0.21,0.49,0.74}
\usepackage[pagebackref,breaklinks,colorlinks,citecolor=cvprblue]{hyperref}
\usepackage[accsupp]{axessibility}

\newcommand{\argmin}{\operatorname*{argmin}}
\newcommand{\D}{\mathcal{D}}
\usepackage{amsmath}
\usepackage{amssymb}
\usepackage{mathtools}
\usepackage{amsthm}
\usepackage[capitalize]{cleveref}

\def\paperID{6} 
\def\confName{CVPR}
\def\confYear{2024}

\title{Calibrating Higher-Order Statistics for Few-Shot Class-Incremental Learning with Pre-trained Vision Transformers}

\author{Dipam Goswami\textsuperscript{1,2} \enspace\enspace Bartłomiej Twardowski\textsuperscript{1,2,3} \enspace\enspace Joost van de Weijer\textsuperscript{1,2}  \\ 
\textsuperscript{1}Department of Computer Science, Universitat Autònoma de Barcelona \\
\textsuperscript{2}Computer Vision Center, Barcelona \quad
\textsuperscript{3}IDEAS-NCBR \\  
{\tt\small \{dgoswami, btwardowski, joost\}@cvc.uab.es}
}

\begin{document}
\maketitle
\begin{abstract}

Few-shot class-incremental learning (FSCIL) aims to adapt the model to new classes from very few data (5 samples) without forgetting the previously learned classes. Recent works in many-shot CIL (MSCIL) (using all available training data) exploited pre-trained models to reduce forgetting and achieve better plasticity. In a similar fashion, we use ViT models pre-trained on large-scale datasets for few-shot settings, which face the critical issue of low plasticity. FSCIL methods start with a many-shot first task to learn a very good feature extractor and then move to the few-shot setting from the second task onwards. While the focus of most recent studies is on how to learn the many-shot first task so that the model generalizes to all future few-shot tasks, we explore in this work how to better model the few-shot data using pre-trained models, irrespective of how the first task is trained. Inspired by recent works in MSCIL, we explore how using higher-order feature statistics can influence the classification of few-shot classes. We identify the main challenge of obtaining a good covariance matrix from few-shot data and propose to calibrate the covariance matrix for new classes based on semantic similarity to the many-shot base classes. Using the calibrated feature statistics in combination with existing methods significantly improves few-shot continual classification on several FSCIL benchmarks. Code is available at \url{https://github.com/dipamgoswami/FSCIL-Calibration}.

\end{abstract}    

\section{Introduction}
\label{sec:intro}
Continual Learning (CL) aims to learn from data in a continuous manner where the data distribution changes over time and the model is expected to not forget the old classes learned in previous tasks, a phenomenon known as catastrophic forgetting~\cite{mccloskey1989catastrophic,robins1995catastrophic,kemker2018measuring}. Class-incremental learning (CIL) refers to learning new classes incrementally over time with the goal of classifying all classes seen so far without any task information~\cite{masana2020class,wang2023comprehensive,van2019three,de2021continual}. While many recent studies focus on many-shot CIL (MSCIL)~\cite{mcdonnell2023ranpac,goswami2023fecam,petit2023fetril,zhu2021prototype} assuming the availability of sufficient training data for all classes, a more challenging setting is few-shot CIL (FSCIL), which considers very few training samples for each class~\cite{tian2024survey,zhou2022forward,zhou2022few,peng2022few,wang2023few,tao2020few,zhang2021few,ahmad2022few}. In this work, we address the FSCIL problem. 

Existing FSCIL methods consider a many-shot first task where a good feature extractor is trained. After the first task, all subsequent tasks are few-shot. The standard practice is to freeze the backbone after the first task and then compute the class-wise prototypes~\cite{vinyals2016matching} by averaging the feature embeddings and classifying using the nearest class mean (NCM) classifier~\cite{rebuffi2017icarl}. While most FSCIL~\cite{zhou2022forward,peng2022few,song2023learning,deng2024expanding,zhou2022few} methods are dependent on the base task learning and focus on how to effectively train the many-shot first task to learn more compact representations of base classes in order to generalize better in future few-shot steps, fewer methods~\cite{wang2023few,goswami2023fecam} propose how to better model the few-shot data in new tasks independent of the first task training. Recently, pre-trained ViTS have been extensively studied for many-shot CIL~\cite{zhou2024continual,mcdonnell2023ranpac,zhang2023slca,goswami2023fecam,zhou2023revisiting}.
Unlike the conventional FSCIL settings using ResNet architectures~\cite{he2016deep} (without pre-training on large-scale datasets), we study how the ViTs~\cite{dosovitskiy2021an} pre-trained on large-scale datasets like ImageNet21k~\cite{ridnik1imagenet} can be exploited for FSCIL.

\begin{figure*}
    \centering
    \begin{subfigure}[b]{0.49\textwidth}
        \centering
        \includegraphics[width=\textwidth]{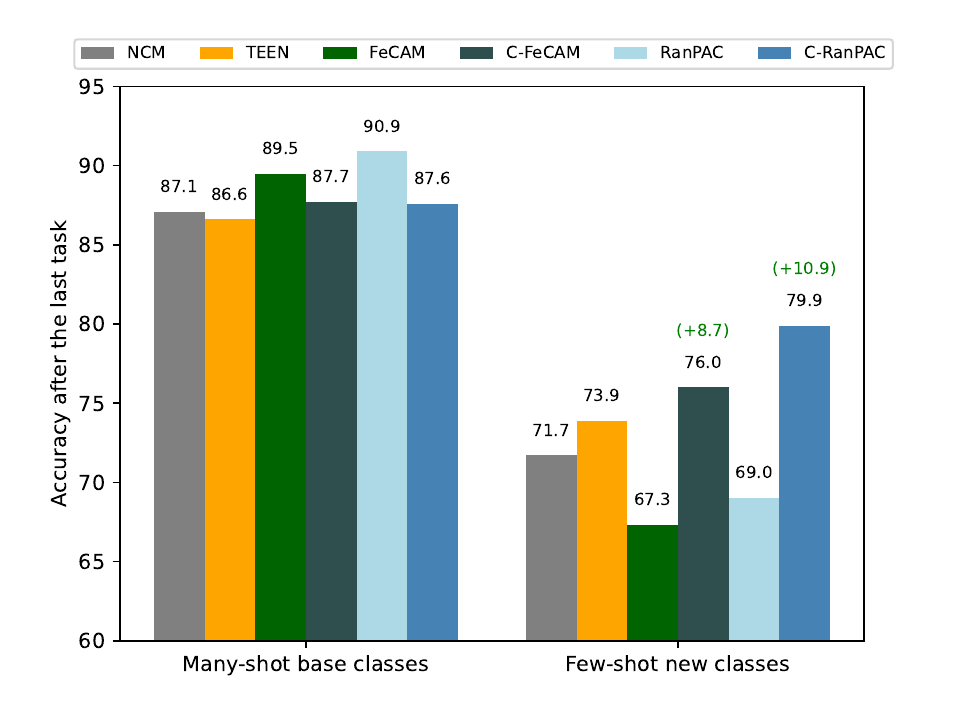}
        \caption{CUB-200 (100 classes in base task, 100 classes in 10 new tasks)}
        \label{fig:sub1}
    \end{subfigure}
    \hfill
    \begin{subfigure}[b]{0.49\textwidth}
        \centering
        \includegraphics[width=\textwidth]{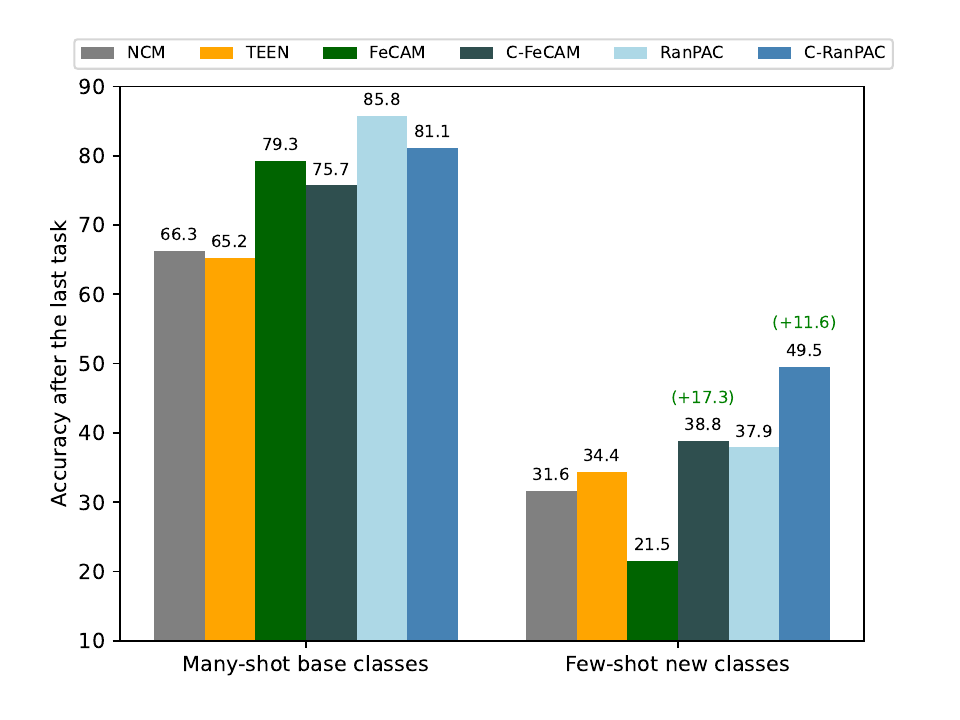}
        \caption{Stanford Cars (98 classes in base task, 98 classes in 7 new tasks)}
        \label{fig:sub2}
    \end{subfigure}
    \caption{Performance of different prototype-based classification methods on FSCIL settings with ViT-B/16 pre-trained on ImageNet-21k. All the methods - NCM, TEEN~\cite{wang2023few}, FeCAM~\cite{goswami2023fecam} and RanPAC~\cite{mcdonnell2023ranpac} are biased towards the base task classes. While TEEN improves the performance on the few-shot classes by prototype calibration, methods using second-order feature statistics - FeCAM and RanPAC performs much poorly on the few-shot classes compared to the many-shot base task classes. This drop in performance for new classes can be attributed to the poor estimates of second-order statistics from few-shot data. We propose to calibrate the covariance matrix of few-shot classes by using the strong covariance estimates of base classes. We observe that on using our proposed calibration, C-FeCAM and C-RanPAC improve performance significantly on the new classes, leading to an overall better accuracy.}
    \label{fig:performance}
\end{figure*}

With the availability of pre-trained weights, a more realistic approach to solving FSCIL would be to focus on how to better use the few-shot data, irrespective of how the base task is adapted. We propose to adapt the pre-trained ViT model to the base classes in the first task with an adaptor~\cite{chen2022adaptformer}, similar to~\cite{zhou2023revisiting}.
After adapting the model in the first task, we study different ways of classifying the test features in the few-shot steps by exploring recent concepts from the many-shot CIL domain which works on a frozen feature space~\cite{goswami2023fecam,mcdonnell2023ranpac}. Recently, FeCAM~\cite{goswami2023fecam} explored the heterogeneous nature of feature distributions in CL settings and proposed to use the anisotropic Mahalanobis distance in the feature space for classification. Another recent work - RanPAC~\cite{mcdonnell2023ranpac} proposed random projections to a very high-dimensional feature space and classified features using a shared Gram matrix instead of the covariance matrix. We use both FeCAM and RanPAC after adapting the pre-trained ViT to the base task and observe poor performance in the few-shot classes (see~\cref{fig:performance}), suggesting that few-shot class statistics are poorly calibrated.

A recent work in FSCIL, TEEN~\cite{wang2023few} proposed to calibrate the prototypes for the new classes based on semantic similarity to the base classes, which resulted in better NCM classification of the new classes. In a similar fashion, we hypothesize that the covariances of new classes can be calibrated based on the covariances of many-shot base classes. This has some evidence in previous works~\cite{petit2023fetril,yangfree,salakhutdinov2012one}. 
We propose to calibrate the covariance matrices of new classes based on the similarity to the base classes. Finally, we use the calibrated statistics with FeCAM and RanPAC and achieve significant improvement over the baseline methods.

We perform experiments on CIFAR100~\cite{krizhevsky2009learning} and fine-grained classification datasets like CUB-200~\cite{wah2011caltech}, Stanford Cars~\cite{krause20133d}, and FGVC-Aircraft~\cite{maji2013fine}, which are very good use cases of few-shot data. We study the commonly used settings with a big first task and also introduce challenging settings with a small first task (equally splitting the dataset into tasks). We demonstrate that calibrated higher-order statistics enable better classification of few-shot classes, which is reflected in the harmonic mean accuracy. The main contributions can be summarized as:
\begin{enumerate}
    \item We explore how knowledge from pre-trained ViTs can be transferred to new few-shot classes in FSCIL settings instead of conventional approaches to training ResNets from scratch.
    \item We identify that higher-order statistics-based classification approaches perform poorly in classifying few-shot classes due to poor estimates of statistics from very few samples.
    \item We propose feature covariance calibration for few-shot classes, exploiting strong covariance estimates of many-shot base classes, thus enabling better classification of few-shot classes using recent state-of-the-art MSCIL methods like FeCAM and RanPAC.
\end{enumerate}

\section{Related Work}
\label{sec:related}

\noindent\textbf{Class-Incremental Learning.} 
Class-Incremental Learning~\cite{masana2020class,wang2023comprehensive,van2019three,de2021continual} learns new classes in incremental tasks and aims to preserve the knowledge of old classes without access to task-ID at inference. While many methods~\cite{douillard2020podnet,dhar2019learning,belouadah2019il2m,hou2019learning,rebuffi2017icarl,liu2023augmented} store samples/exemplars from old classes in memory, recent methods~\cite{petit2023fetril,goswami2023fecam,mcdonnell2023ranpac,zhu2021prototype} propose exemplar-free solutions to CIL. The main challenge of exemplar-free CIL is to prevent forgetting of old classes since it is difficult to distinguish classes from different tasks~\cite{soutif2021importance}. CIL methods typically focus on either regularization approaches~\cite{kirkpatrick2017overcoming,douillard2020podnet,li2017learning,liu2018rotate}. Several methods~\cite{hayes2020lifelong,de2021continualP,janson2022simple,goswami2023fecam,mcdonnell2023ranpac,panos2023session} show that using prototypes and higher-order statistics for classification approaches is very efficient. In this work, we explore how these state-of-the-art MSCIL methods perform in FSCIL settings.

\noindent \textbf{Few-Shot Class-Incremental Learning.}
A common practice in existing FSCIL methods~\cite{tian2024survey,zhou2022forward,peng2022few,song2023learning,deng2024expanding,zhou2022few,hersche2022constrained} is to use a simple NCM classifier based on the class-wise prototypes on the frozen model (where the model is only trained on the first task). FACT~\cite{zhou2022forward} reserves space in the embedding space during the base task training for future classes by allocating virtual prototypes. 
To enable better model generalization in future few-shot tasks, during the base task training 
SAVC~\cite{song2023learning} proposes using semantic-aware fantasy classes, ALICE~\cite{peng2022few} uses angular penalty loss, LIMIT~\cite{zhou2022few} uses fake incremental tasks and EHS~\cite{deng2024expanding} uses expanding the hyperspherical space. Some methods update the network in the few-shot tasks by learning a neural gas network~\cite{tao2020few}, learning soft masks~\cite{kang2022soft}, by weight space rotation process~\cite{kim2022warping}, by self-supervised stochastic classifiers~\cite{kalla2022s3c} or by training a two-branch network with class-aware bilateral distillation~\cite{zhao2023few}. Different from these methods, inspired by neural collapse, NC-FSCIL~\cite{yang2022neural} fixes the prototype positions in a simplex equiangular tight frame and trains a model to map the features to their corresponding prototypes. While all these works use resnet architectures, pre-trained ViTs have been recently used~\cite{tian2024pl} for FSCIL.

Recent methods~\cite{akyureksubspace,wang2023few} propose to calibrate new-class prototypes using semantic information of base classes. Aky{\"u}rek et al.~\cite{akyureksubspace} propose a semantic subspace regularization-based objective to calibrate new class prototypes. Wang et al.~\cite{wang2023few} observe that while the base class prototypes are well-calibrated due to abundant training samples in the many-shot first task, the new class prototypes are biased, resulting in high false-positive classifications of new classes to their most similar base classes. Based on this observation, they propose to simply calibrate the new class prototypes by simply fusing them with the base class prototypes, which are weighted based on their similarity. Motivated by these prototype calibration methods, we propose to extend calibration to include higher-order statistics of few-shot classes.

\section{Method}
\label{sec:method}

The main objective of FSCIL methods is to improve the classification of few-shot classes while not forgetting the base classes. The most commonly studied classification method for FSCIL is to perform NCM and assign the test sample to the closest prototype mean in the feature space using euclidean distance.
We observe that while calibrated prototypes using TEEN~\cite{wang2023few} improve performance marginally compared to naive NCM, the higher-order statistics (covariance matrix or Gram matrix) for new classes are still poorly estimated from a few samples and thus affect classification with methods like FeCAM and RanPAC. We show in~\cref{fig:performance} that directly applying these methods to FSCIL settings results in poor performance on few-shot classes. The main contribution of our work is to show how to get better estimates of the feature distributions of the few-shot classes by calibrating their covariances based on the base class covariances.



\begin{figure}[t]
\centering
\includegraphics[width=\linewidth]{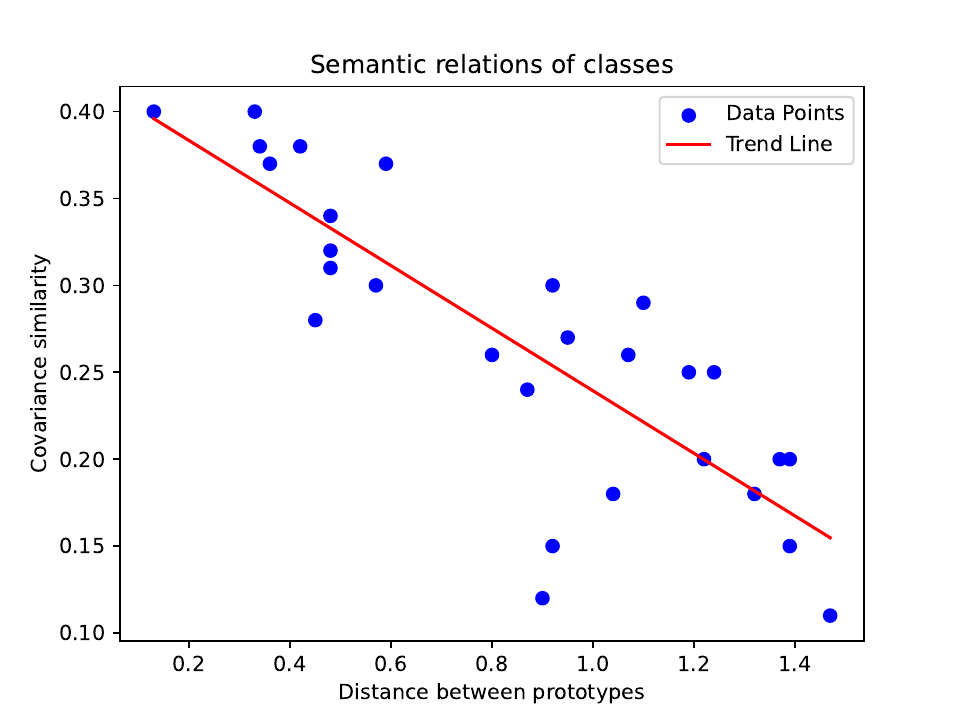}
\caption{Illustration to demonstrate how the similarity of the covariance matrices of classes vary based on the distance between the class prototypes. We train the model on 28 base classes on the Stanford Cars dataset, and plot the covariance similarity with respect to the prototype distance of a new class (from task 1) with the base classes. We observe that the classes with similar prototypes (lesser distance between the prototypes) have higher covariance similarities. } 
\label{fig:semantic}
\end{figure}

We explore how these various methods work with few-shot data when using ViT models pre-trained on large-scale datasets like ImageNet-21k. While pre-trained models have been found to provide very good classification accuracy without any training on the downstream datasets~\cite{janson2022simple,goswami2023fecam}, it is important to adapt the model to the downstream datasets~\cite{zhou2023revisiting,mcdonnell2023ranpac}. Conventionally, FSCIL methods are evaluated with ResNet models (without large-scale pretraining). However, it is more intuitive to use pre-trained knowledge in FSCIL settings due to the availability of very few training samples. In this work, we adapt the pre-trained ViT model on the dataset in the many-shot base task by optimizing the adaptor, similar to~\cite{chen2022adaptformer} and then use the prototype-based methods for classification in the new tasks without any further adaptation.


\subsection{Motivation}
Prototype calibration in TEEN~\cite{wang2023few} aims to exploit the semantic relationships between the base classes on which the model is adapted and the new classes, which have very few samples. The relationships between the means and variances of different classes have been explored in few-shot learning~\cite{yangfree,salakhutdinov2012one}. In few-shot learning~\cite{wang2020generalizing}, class distributions have been calibrated in different forms by simply averaging from few most similar base classes while discarding the current class covariance matrices~\cite{yangfree} or by iteratively learning the distribution calibration~\cite{liu2023learnable}. 
Following the recent success of using higher-order feature statistics for classification~\cite{goswami2023fecam,mcdonnell2023ranpac}, we explore how the covariance matrix of each new class can be calibrated based on their semantic relations to all base classes.

We analyze in~\cref{fig:semantic} how the covariance matrices vary for classes with varying prototype similarities. We observe that the covariances are similar for classes with higher cosine similarity between their prototypes. Thus, we propose to exploit these semantic relationships to calibrate the covariance matrices of new few-shot classes using the strong covariance estimates of many-shot base classes, weighted based on their prototype similarities.


\subsection{Statistics Calibration}

The similarity between a pair of base class prototype $\mu_b$ and a new class prototype $\mu_n$ can be used to compute the weights for averaging all the old class statistics with new ones. We obtain the cosine similarity $S_{b,n}$ between $\mu_b$ and $\mu_n$ as follows:
\begin{equation}
    S_{b,n} = \frac{\mu_{b} \cdot \mu_{n}}{\|\mu_{b}\| \cdot \|\mu_{n}\|} \cdot \tau
\end{equation}
where $\tau$ controls the sharpness of the weight's distribution. Following~\cite{wang2023few}, we consider $\tau = 16$ in our experiments.

The weightage of a new class prototype $\mu_{n}$ corresponding to a base class prototype $\mu_{b}$ can be obtained by performing softmax over all the base class prototypes as follows:
\begin{equation}
w_{b,n} = \frac{e^{S_{b,n}}}{\sum_{i=1}^{B} e^{S_{i,n}}}
\label{softmax-weight-eq}
\end{equation}
such that $\sum_{b=1}^{B} w_{b,n} = 1$ for a new class $n$, where $B$ is the number of base classes.

\noindent\textbf{Prototype Calibration.}
Similar to~\cite{wang2023few}, the biased prototype means of new classes $\mu_n$ can be calibrated as follows:
\begin{equation}
    \hat{\mu}_{n} =  \alpha \, \mu_{n} + (1-\alpha) \, \sum_{b=1}^{B} w_{b,n} \mu_{b}
    \label{proto-calibration}
\end{equation}
where $\alpha$ controls the degree of calibration.

\noindent\textbf{Covariance Calibration.} 
Similar to~\cref{proto-calibration}, we propose to use the softmaxed similarity weights from~\cref{softmax-weight-eq} to calibrate the new class covariances $\Sigma_{n}$ by averaging with the base class covariances $\Sigma_{b}$ as follows:
\begin{equation}
    \hat{\Sigma}_{n} =  \beta \, (\Sigma_{n} + \sum_{b=1}^{B} w_{b,n}\Sigma_{b})   
    \label{cov-calibration}
\end{equation}
where $\beta$ controls the scaling of the covariance matrix.

\subsection{Calibration with existing methods}

\subsubsection{FeCAM}
Goswami et al.~\cite{goswami2023fecam} propose to use a Mahalanobis distance-based classifier to classify feature embeddings at test time. They demonstrate that the anisotropic Mahalanobis distance is more effective than the commonly used Euclidean distance in CL settings for classification in the embedding space. To compute the Mahalanobis distance, FeCAM uses the covariance matrix of the feature embeddings for all classes. In the few-shot setting, the covariance matrix has to be obtained using only 5 samples per class resulting in very poor estimates. To get better estimates of the few-shot class distributions, we propose to use FeCAM with calibrated statistics.

\vspace{5pt}
\noindent\textbf{Calibrated FeCAM.} In order to complement FeCAM for FSCIL, we use calibrated prototypes and covariances for each new class. Similar to~\cite{mcdonnell2023ranpac,goswami2023fecam}, we perform covariance shrinkage to obtain an invertible full-rank covariance matrix.
To account for the shift in feature distributions between base classes on which the backbone is trained and new classes that are not used for training, we follow~\cite{goswami2023fecam} and obtain the correlation matrix from the shrunk covariance matrix by performing correlation normalization.

We compute the Mahalanobis distance between the prototypes and the test features by using the correlation matrix of each class as follows:
\begin{equation}
\label{eq:maha_ncm}
\D_M(f(x),\hat{\mu}_y) = ({f(x)}-\hat{\mu}_y)^T (N{({\hat{\Sigma}_y}+ \gamma I)})^{-1} ({f(x)}-{\hat{\mu}_y})
\end{equation}
where $\hat{\mu}_y$ refers to the calibrated prototypes, $\hat{\Sigma}_y$ refers to the calibrated covariances obtained using~\cref{cov-calibration}, $f(x)$ refers to the features obtained using the test samples, and $N$ denotes the correlation normalization. For the base classes, we apply $\hat{\Sigma}_y = \Sigma_y$.

The features are then classified based on the Mahalanobis distance as follows:
\begin{equation}
y^\ast = \argmin\limits_{y=1,\dots,Y}  \D_M(f(x),\hat{\mu}_y)   
\end{equation}

\begin{table*}[t]
\begin{center}
\resizebox{0.99\textwidth}{!}{
\begin{tabular}{clccccccc|ccccccc}
\toprule
    & & \multicolumn{7}{c}{Small First Task (10 tasks)} & \multicolumn{7}{c}{Big First Task (11 tasks)}              \\
 \cmidrule(lr){3-9} \cmidrule(lr){10-16}
      &    & \multicolumn{5}{c}{$A_{HM}$} & $A_{last}$ & $A_{inc}$ & \multicolumn{5}{c}{$A_{HM}$} & $A_{last}$ & $A_{inc}$  \\
      & Task      & 1 & 3 & 5 & 7 & 9 & & & 2 & 4 & 6 & 8 & 10 &  &  \\
           \midrule
\rule{0pt}{2ex} 
\multirow{6}{*}{\rotatebox[origin=c]{90}{\textbf{CUB200}}}    &     

NCM & 88.48 & 80.95 & 77.75 & 72.19 & 66.23 & 70.25 & 79.68 & 75.9 & 84.21 & 82.16 & 78.35 & 71.33 & 79.43 & 84.24 \\
 & TEEN~\cite{wang2023few} & 89.77 & 81.33 & 78.57 & 71.76 & 67.25 & 70.52 & 80.04 & 78.01 & 85.34 & 82.89 & 79.44 & 73.06 & 80.24 & 84.83 \\
 & FeCAM~\cite{goswami2023fecam} & 91.14 & 80.33 & 76.59 & 70.55 & 65.24 & 70.25 & 79.93 & 71.05 & 82.0 & 77.65 & 74.36 & 64.42 & 79.43 & 84.03 \\
 & C-FeCAM & 92.45 & 82.94 & 79.29 & 74.01 & 69.45 & 72.68 & 81.87 & 80.56 & 86.59 & 83.36 & 81.59 & \underline{74.35} & \underline{81.86} & \underline{86.12} \\
 & RanPAC~\cite{mcdonnell2023ranpac} & 91.87 & 83.54 & 79.23 & 75.11 & \underline{70.23} & \underline{74.65} & \underline{82.91} & 70.94 & 81.65 & 80.76 & 80.12 & 67.72 & 79.96 & 84.87 \\
 & C-RanPAC & 92.09 & 84.16 & 81.47 & 76.72 & \textbf{72.77} & \textbf{76.22} & \textbf{83.73} & 84.63 & 86.47 & 85.30 & 85.07 & \textbf{78.23} & \textbf{83.72} & \textbf{87.43} \\
 \bottomrule

\rule{0pt}{3ex}    
\multirow{6}{*}{\rotatebox[origin=c]{90}{\textbf{FGVC-Aircraft}}}    &       
NCM & 27.21 & 20.85 & 17.1 & 13.07 & 15.01 & 13.98 & 21.37 & 24.31 & 5.6 & 16.88 & 18.86 & 25.75 & 27.45 & 33.93 \\
 & TEEN~\cite{wang2023few} & 29.31 & 21.65 & 18.35 & 14.04 & 15.02 & 14.16 & 21.82 & 26.02 & 5.6 & 18.71 & 19.39 & 27.67 & 27.63 & 33.95 \\
 & FeCAM~\cite{goswami2023fecam} & 25.85 & 23.09 & 21.33 & 14.29 & 16.38 & 15.42 & 23.54 & 19.08 & 1.2 & 10.32 & 12.59 & 22.58 & 29.49 & 37.87 \\
 & C-FeCAM & 32.42 & 26.39 & 23.63 & 15.71 & 17.55 & 16.38 & 24.9 & 30.19 & 15.81 & 20.78 & 24.96 & 29.42 & 31.17 & 39.13 \\
 & RanPAC~\cite{mcdonnell2023ranpac} & 33.37 & 27.78 & 25.88 & 17.21 & \underline{21.89} & \underline{21.12} & \underline{32.0} & 26.15 & 6.81 & 21.32 & 14.84 & \underline{35.94} & \underline{38.22} & \underline{48.53} \\
 & C-RanPAC & 41.61 & 32.54 & 28.79 & 19.97 & \textbf{24.33} & \textbf{21.87} & \textbf{33.63} & 43.26 & 20.61 & 29.77 & 25.52 & \textbf{43.89} & \textbf{40.32} & \textbf{50.43} \\

\bottomrule

\end{tabular}
}
\caption{Evaluation of methods in FSCIL settings on CUB200 and FGVC-Aircraft datasets with small and big first task settings. Best results in \textbf{bold} and second-best results are \underline{underlined}.}
\label{table1}
\end{center}
\end{table*}

\subsubsection{RanPAC}
McDonnell et al.~\cite{mcdonnell2023ranpac} randomly project the features using non-linear activations to a very high dimensional space and perform classification in high dimensions where the linear separability of features are better. They advocated using the Gram matrix of features for prototype-based classification due to reduced off-diagonal correlations among the class prototypes, which enables better separability of classes. RanPAC proposed to compute the probability scores as follows:
\begin{equation}
    S_y = \phi(f(x)W)(G+\lambda I)^{-1}c_y.
\end{equation}
where $W$ refers to the random projection weights which are kept frozen after the first task, $\phi(.)$ denotes element-wise non-linear activation function, $G$ refers to the summation of gram matrices of all classes, $\lambda$ refers to the shrinkage parameter and $c_y$ refers to the summation of features of class $y$.
The Gram matrix $G$ and the class prototypes (without averaging) $c_y$ are updated in every task. 

\vspace{5pt}
\noindent\textbf{Calibrated RanPAC.}
In the FSCIL scenario, the estimates of $G$ and $c_y$ are not good and biased for the new few-shot classes, which results in high scores for the base classes and poor classification for the new classes. We propose to use the calibrated means and covariances (from~\cref{cov-calibration}) to assume gaussian distributions of new classes and then sample features from these calibrated distributions as follows:
\begin{equation}
\label{eq:sample}
    \hat{f}(x) \sim \mathcal{N}(\hat{\mu}_y, \hat{\Sigma}_y)
\end{equation}

We compute $G$ and $c_y$ from these sampled features as follows:
\begin{equation}
\label{eq:gram}
    G = \sum_{t=1}^{T}\sum_{i=1}^{N_t} h_{t,n} \cdot h_{t,n}, \hspace{10pt} C = \sum_{t=1}^{T}\sum_{i=1}^{N_t} h_{t,n} \cdot y_{t,n}
\end{equation}
where $h_{t,n} = \phi(\hat{f}(x)W)$ and $(\cdot)$ refers to the outer products. Thus, we use the calibrated feature distributions of new few-shot classes and perform classification in the randomly projected high dimensional space.
\section{Experiments}

\noindent \textbf{Datasets.} We perform experiments on several publicly available datasets. CIFAR100~\cite{krizhevsky2009learning} contains 50,000 training images and 10000 images for testing, divided among 100 classes. CUB-200~\cite{wah2011caltech} contains 5994 training images, 5794 testing images and covers 200 classes of birds. Stanford Cars~\cite{krause20133d} consists of 196 classes of car models with 8144 images for both training and testing. FGVC-Aircraft~\cite{maji2013fine} has 10,200 images covering 102 classes of aircraft variants, out of which we use randomly selected 100 classes. 

\noindent \textbf{Settings.} We use all available samples for each class in the many-shot base task and only 5 samples from each class in the few-shot tasks. We follow both the commonly used setting of having a big first task (50\% of the classes in the first task) as well as the challenging small start setting (equally split the dataset in all tasks). In the big start setting, we use 50 classes in the base task for the FGVC-Aircraft dataset, 100 classes for CUB-200 and 98 classes for Stanford Cars. For the small start setting, we use 10 classes in the base task and all subsequent tasks for CIFAR-100 and FGVC-Aircraft datasets, 20 classes for CUB-200 and 28 classes for Stanford Cars in all tasks.

\begin{table*}[t]
\begin{center}
\resizebox{0.9\textwidth}{!}{
\begin{tabular}{lcccccc|cccccc}
\toprule
    & \multicolumn{6}{c}{Small First Task (7 tasks)} & \multicolumn{6}{c}{Big First Task (8 tasks)}              \\
 \cmidrule(lr){2-7} \cmidrule(lr){8-13}
        & \multicolumn{4}{c}{$A_{HM}$} & $A_{last}$ & $A_{inc}$ & \multicolumn{4}{c}{$A_{HM}$} & $A_{last}$ & $A_{inc}$  \\
       Task    & 1 & 2 & 4 & 6 & & & 1 & 3 & 5 & 7 &  &  \\
           \midrule
NCM & 44.15 & 35.66 & 24.22 & 23.59 & 25.71 & 39.52 & 47.72 & 41.17 & 41.64 & 41.48 & 48.91 & 57.74 \\
 TEEN~\cite{wang2023few} & 45.66 & 37.57 & 25.49 & 24.57 & 26.46 & 40.20 & 51.48 & 45.15 & 44.14 & 43.47 & 49.83 & 58.49 \\
 FeCAM~\cite{goswami2023fecam} & 41.71 & 32.63 & 24.22 & 23.4 & 27.07 & 41.61 & 39.12 & 33.69 & 34.37 & 30.95 & 50.39 & 62.43 \\
 C-FeCAM & 54.18 & 43.61 & 31.87 & 28.86 & 30.71 & 45.65 & 57.91 & 52.15 & 50.03 & \underline{49.16} & 57.21 & 66.77 \\
 RanPAC~\cite{mcdonnell2023ranpac} & 58.58 & 48.14 & 39.13 & \underline{35.65} & \underline{38.3} & \underline{54.15} & 53.22 & 54.62 & 48.13 & 47.61 & \underline{61.85} & \underline{72.12} \\
 C-RanPAC & 62.63 & 51.97 & 43.64 & \textbf{38.18} & \textbf{40.73} & \textbf{56.05} & 65.98 & 62.24 & 58.78 & \textbf{57.82} & \textbf{65.3} & \textbf{74.32} \\
\bottomrule

\end{tabular}
}
\caption{Evaluation of methods in FSCIL settings on Stanford Cars dataset with small and big first task settings. Best results in \textbf{bold} and second-best results are \underline{underlined}.}
\label{table2}
\end{center}
\end{table*}

\noindent \textbf{Implementation Details.} Similar to~\cite{zhou2023revisiting}, we use the ViT-B/16 model from the timm library which is pre-trained on ImageNet-21K and then finetuned on ImageNet-1k. We follow~\cite{zhou2023revisiting,chen2022adaptformer} and use a ViT adaptor for adapting the pretrained ViT to the dataset in the first task for 40 epochs. We adapt the model in the base task and use the same model weights for different methods to ensure fair comparison.  We follow the code-base from~\cite{sun2023pilot}. Different from~\cite{zhou2023revisiting}, we do not merge the embeddings obtained using the old model with the current model embeddings and use only the current ones. For the prototype calibration in TEEN, C-FeCAM, and C-RanPAC, we use $\alpha = 0.9$ for CUB-200, Stanford Cars, and FGVC-Aircraft and $\alpha=0.75$ for CIFAR-100. For the covariance calibration, we use $\beta = 1$ for C-FeCAM and $\beta = 0.5$ for C-RanPAC for all datasets. For C-RanPAC, we sample 800 points in the feature space for each class using~\cref{eq:sample} to compute the matrix $G$ and class-wise prototypes $C$ in~\cref{eq:gram}.

For FeCAM, we use $\gamma = 100$ following the original implementation from~\cite{goswami2023fecam}. We do not use the tukey's transformation from~\cite{goswami2023fecam} since the pre-trained ViTs do not have a final ReLU activation layer and can have negative values in the feature representations. 
For RanPAC, we follow the original implementation~\cite{mcdonnell2023ranpac} and randomly project the features to 10,000 dimensional space. Following~\cite{mcdonnell2023ranpac}, we optimize the shrinkage parameter by computing $G$ and $c_y$ on 80\% of the training set for some value of $\lambda$ and then selecting the value which leads to minimum mean squared error between the predictions and the labels on the other 20\% of the training set. However, since the training set is very small due to the few-shot nature of new tasks, the optimization of $\lambda$ in new tasks leads to instability and performance collapse. In order to avoid this collapse, we optimize the $\lambda$ value in the first task and then fix it for all tasks.

\begin{table}[t]
\begin{center}
\resizebox{\linewidth}{!}{
\begin{tabular}{lccccccc}
\toprule
       & \multicolumn{5}{c}{$A_{HM}$} & $A_{last}$ & $A_{inc}$ \\
     Task      & 1 & 3 & 5 & 7 & 9 & &  \\
           \midrule
NCM & 92.26 & 88.42 & 81.92 & 76.79 & 80.18 & 79.46 & 86.57  \\
TEEN~\cite{wang2023few} & 93.45 & 89.44 & 81.71 & 75.83 & 80.95 & 79.39 & 86.95 \\
FeCAM~\cite{goswami2023fecam} & 92.19 & 87.99 & 79.37 & 74.26 & 78.94 & 78.42 & 86.04 \\
C-FeCAM & 94.16 & 89.63 & 81.23 & 76.11 & \underline{81.31} & \underline{79.73} & \underline{87.24} \\
RanPAC~\cite{mcdonnell2023ranpac} & 91.79 & 85.25 & 80.24 & 75.37 & 79.85 & 78.96 & 85.73 \\
C-RanPAC & 92.51 & 88.89 & 82.88 & 77.49 & \textbf{81.84} & \textbf{81.32} & \textbf{87.55} \\
\bottomrule
\end{tabular}
}
\caption{Evaluation of methods in FSCIL settings on CIFAR100 dataset (small-start setting with 10 tasks). Best results in \textbf{bold} and second-best results are \underline{underlined}.}
\label{table3}
\end{center}
\end{table}

Note that FeCAM stores one ($768\times768$) covariance matrix per class and RanPAC stores one big shared Gram matrix ($10000\times10000$) across tasks in addition to the class prototypes. For the proposed calibration, we need to additionally store the covariance matrices of only the base classes of ($768\times768$) size, to use them for calibration in future tasks.

\noindent \textbf{Evaluation.} We evaluate the performance of NCM~\cite{rebuffi2017icarl,snell2017prototypical,janson2022simple}, TEEN~\cite{wang2023few}, FeCAM~\cite{goswami2023fecam} and RanPAC~\cite{mcdonnell2023ranpac} with the proposed statistics calibration. We refer to the calibrated versions as C-FeCAM and C-RanPAC. We evaluate in terms of the average accuracy after the last task ($A_{last}$), the average of the incremental accuracy from all tasks ($A_{inc}$) and the harmonic mean accuracy ($A_{HM}$) after each task. The harmonic mean accuracy better reflects the stability-plasticity trade-off. It is computed as 
\begin{equation}
    A_{HM} = \frac{2 A_{old} A_{new}}{A_{old} + A_{new}}  
\end{equation}
Where $A_{old}$ refers to the accuracy of all classes seen before the current task and $A_{new}$ refers to the accuracy of the current task classes. The $A_{HM}$ is low if the current task performance is poor, irrespective of the performance of the old classes. Thus, it is a good metric for evaluation in FSCIL benchmarks and is used in previous works~\cite{wang2023few,peng2022few}. We show the $A_{HM}$ after multiple tasks which reflects on the improved performance of the few-shot classses.

\subsection{Quantitative Evaluation} 
We evaluate how the different methods discussed above perform in different settings in~\cref{table1,table2,table3}. For each setting, we report the $A_{HM}$ after alternate tasks, the average accuracy after the last task and the average incremental accuracy. We observe that TEEN~\cite{wang2023few} improves the harmonic mean accuracy marginally compared to NCM across all settings. While FeCAM shows similar performance to TEEN, it has a lower harmonic mean accuracy (64.42 compared to 71.33 in NCM for big-start CUB-200) due to poor estimates of covariance matrix for new classes. On using the proposed calibration, C-FeCAM improves significantly on the last task $A_{HM}$ (from 64.42 to 74.35 for big-start CUB-200). Similarly, RanPAC has a lower last task $A_{HM}$ (67.72 compared to 71.33 in NCM for big-start CUB-200) and on calibration, we see that C-RanPAC improves over RanPAC by 10.51 percentage points (pp).

\begin{figure*}
    \centering
    \begin{subfigure}[b]{0.33\textwidth}
        \centering
        \includegraphics[width=\textwidth]{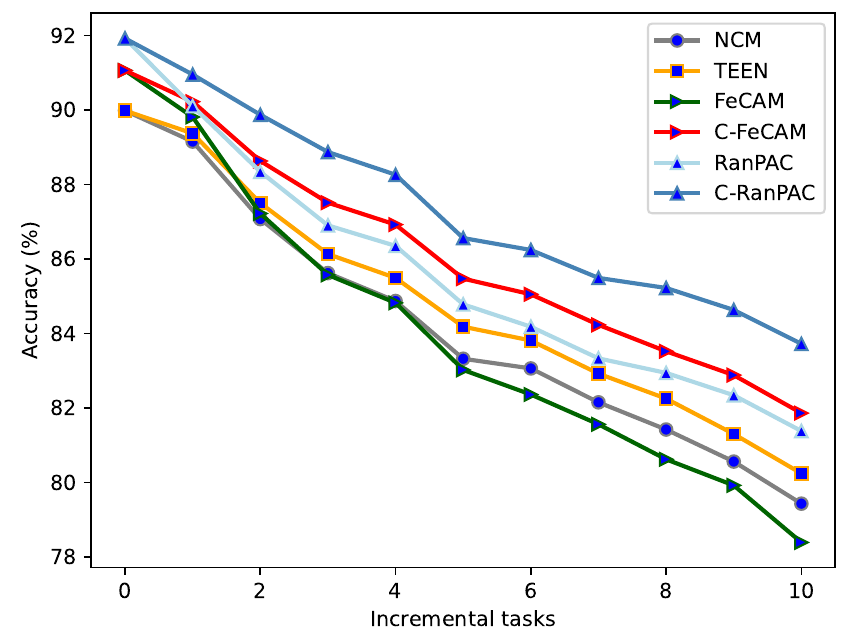}
        \caption{CUB-200}
        \label{fig:sub1}
    \end{subfigure}
    \hfill
    \begin{subfigure}[b]{0.32\textwidth}
        \centering
        \includegraphics[width=\textwidth]{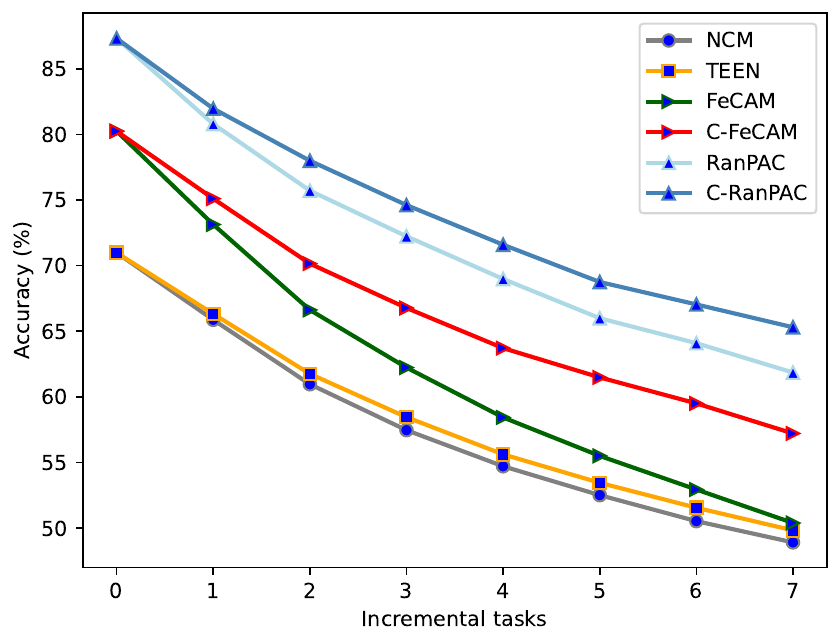}
        \caption{Stanford Cars}
        \label{fig:sub2}
    \end{subfigure}
    \hfill
    \begin{subfigure}[b]{0.32\textwidth}
        \centering
        \includegraphics[width=\textwidth]{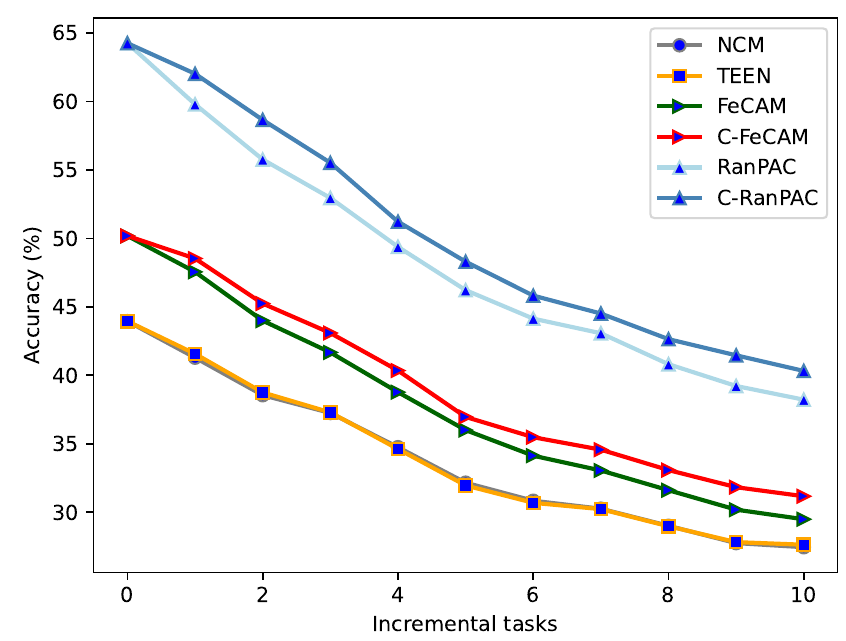}
        \caption{FGVC-Aircraft}
        \label{fig:sub2}
    \end{subfigure}
    \caption{Accuracy after each incremental task for big-start settings on CUB-200, Stanford Cars and FGVC-Aircraft. Our proposed statistics calibration improves the average accuracy consistently after all tasks.}
    \label{fig:curves}
\end{figure*}

\begin{table}[t]
\begin{center}
\resizebox{0.9\linewidth}{!}{
\begin{tabular}{lccc|ccc}
\toprule
    & \multicolumn{3}{c}{C-FeCAM} & \multicolumn{3}{c}{C-RanPAC}              \\
 \cmidrule(lr){2-4} \cmidrule(lr){5-7} 
  $\beta$ & $A_{HM}$ & $A_{last}$ & $A_{inc}$ & $A_{HM}$ & $A_{last}$ & $A_{inc}$ \\
       \midrule
0.5 & 27.4 & 30.05 & 44.57 & \textbf{38.18} & \textbf{40.73} & \textbf{56.05} \\
1.0 & \textbf{28.86} & \textbf{30.71} & \textbf{45.65} & 37.21 & 40.0 & 55.45 \\
1.5 & 28.81 & 30.26 & 45.55 & 35.55 & 38.09 & 54.04 \\
\bottomrule
\end{tabular}
}
\caption{Ablation on the impact of $\beta$ in~\cref{cov-calibration} for Stanford Cars (small-start setting). Best results in \textbf{bold}.}
\label{table5}
\end{center}
\end{table}

\begin{table}[t]
\begin{center}
\resizebox{\linewidth}{!}{
\begin{tabular}{cccccc}
\toprule
 & $\mu_n$~\cref{proto-calibration} & $\Sigma_n$~\cref{cov-calibration} & $A_{HM}$ & $A_{last}$ & $A_{inc}$ \\
 \midrule 
FeCAM & $\times$ & $\times$ &  23.4 & 27.07 & 41.61  \\
C-FeCAM & $\checkmark$ & $\times$ & 24.45 & 28.16 & 42.65 \\
C-FeCAM & $\checkmark$ & $\checkmark$ & \textbf{28.86} & \textbf{30.71} & \textbf{45.65} \\
       \midrule
RanPAC & $\times$ & $\times$ & 35.65 & 38.3 & 54.15 \\
RanPAC* & $\times$ & $\times$ & 36.13 & 39.63 & 55.45 \\
C-RanPAC & $\checkmark$ & $\times$ & 37.32 & 39.85 & 55.39 \\
C-RanPAC & $\checkmark$ & $\checkmark$ & \textbf{38.18} & \textbf{40.73} & \textbf{56.05} \\
\bottomrule
\end{tabular}
}
\caption{Ablation on the impact of prototype calibration and covariance calibration for Stanford Cars (small-start setting). Best results in \textbf{bold}.}
\label{table6}
\end{center}
\end{table}

Having a big first task helps to adapt the model better to the dataset and thus the big first task settings show better performance compared to the small first task. Also, for the proposed statistics calibration, having more base classes will help to better calibrate the new few-shot class statistics. We observe this trend for all settings with higher margin of improvement in big-start settings. Using the challenging FGVC-Aircraft dataset in~\cref{table1}, we notice that all the methods struggle more in the small-start settings but still both C-FeCAM and C-RanPAC improve the last task $A_{HM}$ by 1.17 pp and by 2.44 pp respectively, over the uncalibrated variants. For the big-start setting on FGVC-Aircraft, we see a bigger improvement of 6.84 pp and 7.95 pp for C-FeCAM and C-RanPAC respectively.

In~\cref{table2}, we observe a similar trend for Stanford Cars, where C-FeCAM improves the last task $A_{HM}$ significantly over FeCAM by 5.46 pp for small-start setting and by 18.21 pp for big-start setting. Similarly, C-RanPAC improves over RanPAC by 2.53 pp and 10.21 pp for small-start and big-start settings respectively. We observe the same trend for CIFAR-100 in~\cref{table3}, where C-FeCAM outperforms FeCAM by 2.37 pp and C-RanPAC outperforms RanPAC by 1.99 pp for last task $A_{HM}$. These quantitative evaluations suggest that simply calibrating the second-order feature statistics by exploiting the base class statistics can improve few-shot classification significantly without any extra training complexity.

\subsection{Ablation Studies}

We perform ablation studies to observe the impact of the covariance scaling factor $\beta$ in~\cref{table5} and also the impact of the prototype calibration and the covariance calibration separately in~\cref{table6}. We observe both prototype calibration and covariance calibration improves the harmonic mean accuracy. The covariance calibration has a significant impact on the improvement of performance using both FeCAM and RanPAC. We also compare with the variant of RanPAC (denoted by RanPAC*) where we sample points from the class distributions and compute the probability instead of using just the available embeddings from few-shot training data. 

Furthermore, to analyze how the average accuracy changes after every task, we show the accuracy plots in~\cref{fig:curves}. We observe that using higher-order feature statistics, FeCAM and RanPAC already achieves better accuracy in the base task. In the incremental tasks, the proposed statistics calibration further improves both the methods and achieves a significant improvement after the last task. While the calibration improves the average accuracy after each task, the main improvement with the calibration is reflected in the harmonic mean accuracy in~\cref{table1,table2,table3} due to significantly better classification of few-shot classes.
\section{Conclusion}
In this work, we explore how prototype- and higher-order statistics-based classification methods work in FSCIL settings when using ViT models pre-trained on large scale datasets. We identify that although the higher-order statistics-based methods like FeCAM and RanPAC performs very well with many-shot data and in existing MSCIL benchmarks~\cite{mcdonnell2023ranpac,goswami2023fecam}, these methods struggles with few-shot data due to poor and biased estimates of distribution statistics (see~\cref{fig:performance}). We propose to perform a simple yet effective statistics calibration by using the strong statistics estimates which are computed for the many-shot base classes. We demonstrate that using the proposed calibration outperforms all existing methods across multiple settings and datasets by a significant margin. We highlight the improvement in the classification of few-shot classes using the harmonic mean accuracy which is consistently better after all tasks with our proposed calibration method.

\paragraph{Acknowledgement.} We acknowledge projects TED2021-132513B-I00 and PID2022-143257NB-I00, financed by MCIN/AEI/10.13039/501100011033 and FSE+ and the Generalitat de Catalunya CERCA Program.
This work was partially funded by the European Union under the Horizon Europe Program (HORIZON-CL4-2022-HUMAN-02) under the project ``ELIAS: European Lighthouse of AI for Sustainability", GA no. 101120237. Bartłomiej Twardowski acknowledges the grant RYC2021-032765-I.

{
    \small
    \bibliographystyle{ieeenat_fullname}
    \bibliography{main}
}

\end{document}